\documentclass{article}
\usepackage[preprint]{neurips_2023}
\usepackage{natbib}

\usepackage[utf8]{inputenc} 
\usepackage[T1]{fontenc}    
\usepackage{hyperref}       
\usepackage{url}            
\usepackage{booktabs}       
\usepackage{amsfonts}       
\usepackage{nicefrac}       
\usepackage{microtype}      
\usepackage{xcolor}         

\usepackage{graphicx}
\usepackage{fixltx2e}
\usepackage{authblk}
\usepackage{xtab}
\usepackage{siunitx}
\usepackage{caption}
\usepackage{subcaption}
\usepackage{multirow}
\usepackage{booktabs}

\begin{document}

\title{Area is all you need : repeatable elements make stronger adversarial attacks}

\author{%
  Dillon Niederhut \\
  Novi Labs \\
  Austin, TX, 78746 \\
  \texttt{dillon.niederhut@gmail.com} \\
}

\maketitle

\begin{abstract}
    Over the last decade, deep neural networks have achieved state of the art in computer vision tasks. These models, however, are susceptible to unusual inputs, known as adversarial examples, that cause them to misclassify or otherwise fail to detect objects. Here, we provide evidence that the increasing success of adversarial attacks is primarily due to increasing their size. We then demonstrate a method for generating the largest possible adversarial patch by building a adversarial pattern out of repeatable elements. This approach achieves a new state of the art in evading detection by YOLOv2 and YOLOv3. Finally, we present an experiment that fails to replicate the prior success of several attacks published in this field, and end with some comments on testing and reproducibility.
\end{abstract}

\section{Introduction}

    Deep neural networks (DNNs) have become the tool of choice for image computer vision (CV). The ascent of neural networks for these tasks is generally recognized as starting with AlexNet, which achieved state of the art on ImageNet \citep{krizhevsky_imagenet_2017}. As neural networks became larger and more complex, the ImageNet benchmark became less useful as a differentiator, so more difficult benchmarks were created \citep{everingham_pascal_2010, lin_microsoft_2015}. New models were developed for these, including YOLO (now up to YOLOv8) and RCNN, followed by FasterRCNN and Detectron \citep{redmon_you_2016,redmon_yolov3_2018,redmon_yolo9000_2016,girshick_rich_2014,ren_faster_2017,wu2019detectron2}.

    These models are susceptible to unusual inputs, known as adversarial examples, which reduce their accuracy. While this was known in classical machine learning, the paper showing this also applied to neural networks was published in 2013 \citep{biggio_wild_2018,szegedy_intriguing_2013,goodfellow_explaining_2014}. Initial examples were formulated as an optimization problem, where the goal was to generate a pattern that minimized accuracy for a specific model on a specific input, while limiting the change in pixel values. The rationale behind the constraint was that the attack would be intercepted if the perturbation was too obvious.

    Later work adopted a threat model where it was not possible to intercept a model input between the sensor and the model, and focused on optimizing attacks across a set of inputs, and under a set of noise-inducing factors likely to occur in physical environments. This led to attacks that were bounded in area but unbounded in pixel value changes. This early work was often deployed in the form of stickers or printed cut-outs that could be applied to or placed over an object \citep{brown_adversarial_2017, thys_fooling_2019, eykholt_robust_2017}.

    This is an important area of research due to the magnitude of both positive and negative outcomes arising from experimentation in avoiding detection, or what are commonly known as evasion attacks. Some have argued, for example, that evasion attacks are an important tool in supporting the right to privacy \citep{das_subversive_nodate}. Conversely, it's possible for evasion strategies themselves to cause negative outcomes, e.g. when they are used to gain unlawful access to sensitive areas.

    Here, we argue that increase in reported attack success rates (ASR) is attributable to increasing the attack area. In other words, it might be true that techniques like adding learned non-linear warping have made these attacks better optimized. It's also possible that the increase in success is determined by area, and that these other methods appear related because the correlate with larger attack areas. Table~\ref{prior_work_table} shows a summary of these results, presenting the best reported metric for each test model, and separating physical tests (R) from virtual tests (V).

    \begin{table}[h]
        \caption{Prior results summary}
        \label{prior_work_table}
        \centering
        \begin{tabular}{c c c c c c c}
            \toprule
            \multicolumn{3}{c}{} & \multicolumn{3}{c}{Reported metric} \\
            Reference & Test & Size & YOLOv2 & YOLOv3 & Faster RCNN \\
            \midrule
            \citep{thys_fooling_2019} & V & Large poster & $recall\sim25\%$ & - & - \\
            \citep{xu_evading_2019} & V & Partial shirt & $ASR\sim79\%$ & - & $ASR\sim65\%$\\
            \citep{xu_evading_2019} & R & Partial shirt & $ASR\sim63\%$ & - & $ASR\sim52\%$\\
            \citep{huang_universal_2020} & V & Partial shirt/pants & - & - & $AP\sim11\%$\\
            \citep{huang_universal_2020} & R & Partial shirt/pants & - & - & $AP\sim19\%$\\
            \citep{wu_making_2020} & R & Whole shirt & $ASR\sim50\%$ & $ASR\sim0\%$ & $ASR\sim0\%$\\
            \citep{wu_making_2020} & R & Large poster & $ASR\sim70\%$ & $ASR\sim40\%$ & $ASR\sim10\%$\\
            \citep{li_invisibilitee_2022} & V & Whole Shirt & - & $AP_{50}\sim2\%$ & $AP_{50}\sim1\%$\\
            \bottomrule
        \end{tabular}
    \end{table}

    Next, we describe a method for generating an adversarial patch that maintains its attack success when printed across an entire obejct. We do this by tiling a very small patch across objects during training and optimizing the loss over the resulting pattern. In our testing, we include two adversarial defenses, and transferability across model architectures.

    Finally, we attempt to replicate adversarial patch success rates reported in prior research, but under our testing conditions. We do this by lifting attacks from published papers, deploying them on physical objects, and applying standard object detection models to the result.

\section{Prior work} \label{prior_work}

    \citet{liu_dpatch_2019} generated attacks by modifying an input that was unbounded in in pixel magnitudes, but was bounded in area. This was an example of an adversarial "patch". \citet{brown_adversarial_2017} extended this threat model to one where the adversary can only change the environment, showing that you could attack CV algorithms in real life.

    \citet{athalye_synthesizing_2017}. observed that physically deployed attacks had lower success than would be expected given the training loss. They explained this by the presence of noise-inducing factors like rotation, lighting, sensor noise, and image compression, and developed expectation over transformations (EOT), a set of data augmentation strategies to overcome this.

    \citet{thys_fooling_2019} showed that an adversarial patch could be used for evasion attacks for humans. This involved placing a large adversarial patch on a poster in front of the experimenter at test time. \citet{eykholt_robust_2017} showed a similar effect for using stickers to cause object detectors to ignore stop signs.

    \citet{xu_evading_2019} demonstrated that it was possible to achieve attack succes by printing a design on a t-shirt. They argued that the key to their success was modeling the deformation of fabric using thin-plate splines (TPS), and then applying this as a data augmentation step.

    \citet{wu_making_2020} argued that these t-shirt attacks transferred well between different object detection models. Their deployment strategy added a reflection of the patch to the shirt sleeves to increase the attack area.

    Invisibili-tee took the idea of an adversarial t-shirt one step further, by placing an attack on an entire shirt, instead of applying a rectangular patch \citep{li_invisibilitee_2022}. They achieved this by projecting a rectangular patch onto a polygon-bounded area around the shirt in simulation.

    \citet{huang_universal_2020} generated adversarial patches that would appear to be depictions of unremarkable things to avoid clothing that looked consipcuous. They did this by including an additional loss that attempted to keep the adversarial patch close to a set of training images.

\section{Experiment 1 - patch area}

    We begin by noting in Table~\ref{prior_work_table} the historic increase in both attack success and adversarial patch size. In experiment 1, we recreate this trend under experimental conditions to evaluate its comparative importance, without the confounds of different training methods.

    \subsection{Methods} \label{experiment_1_methods}

        \begin{figure}
            \centering
            \begin{subfigure}[b]{0.12\linewidth}
                \centering
                \includegraphics[width=\linewidth]{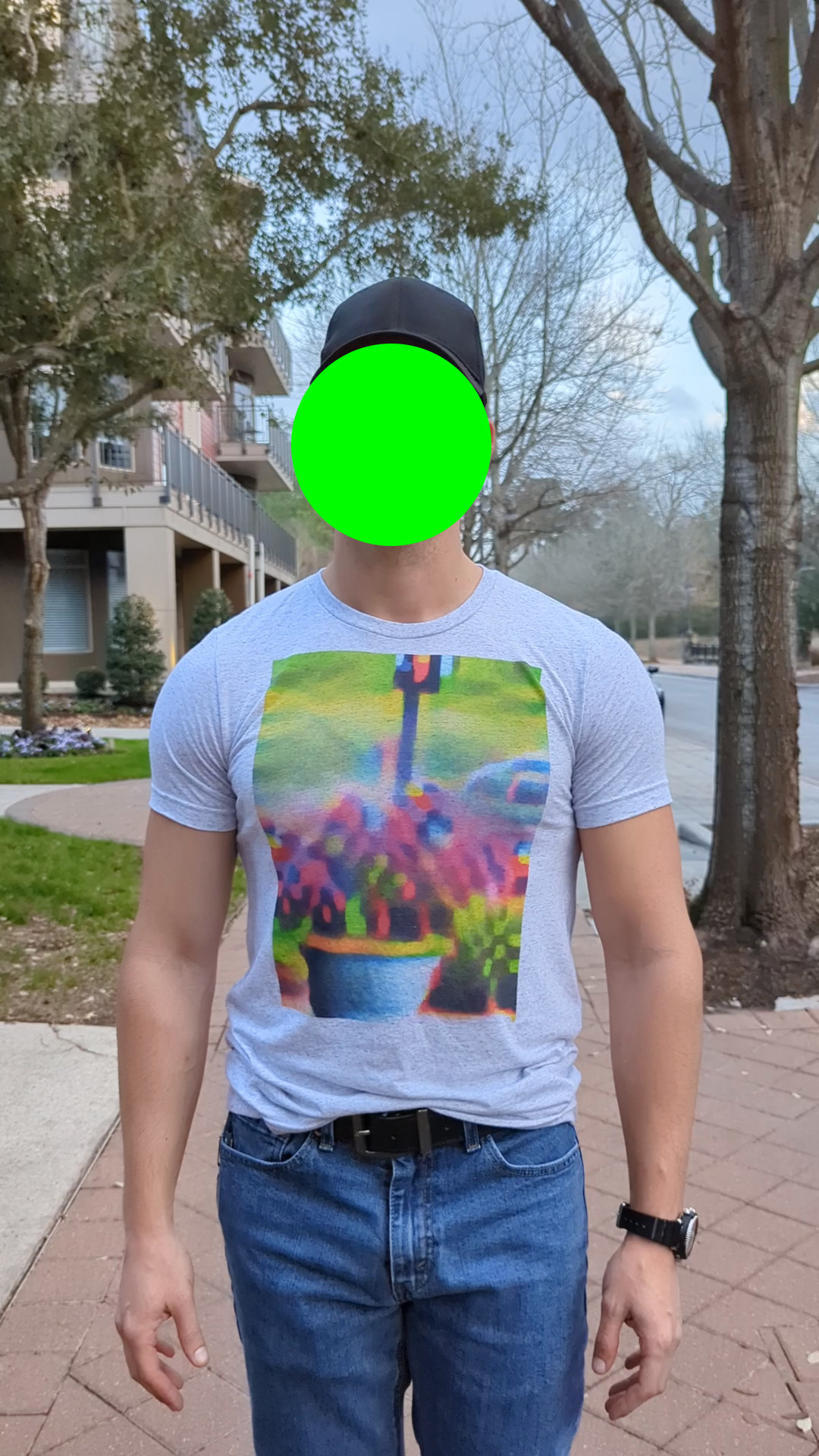}
                \caption{}
            \end{subfigure}
            \begin{subfigure}[b]{0.12\linewidth}
                \centering
                \includegraphics[width=\linewidth]{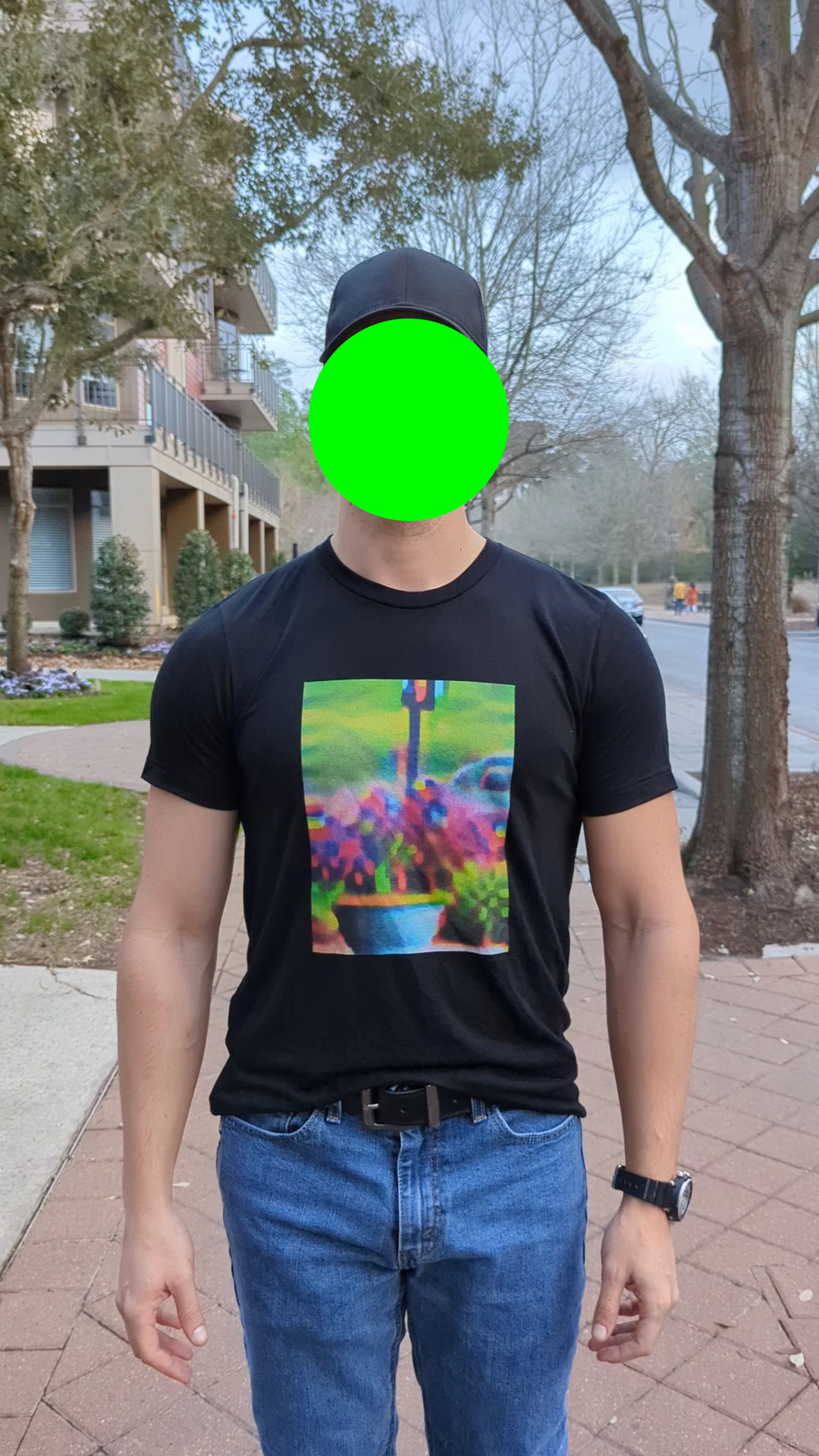}
                \caption{}
            \end{subfigure}
            \begin{subfigure}[b]{0.12\linewidth}
                \centering
                \includegraphics[width=\linewidth]{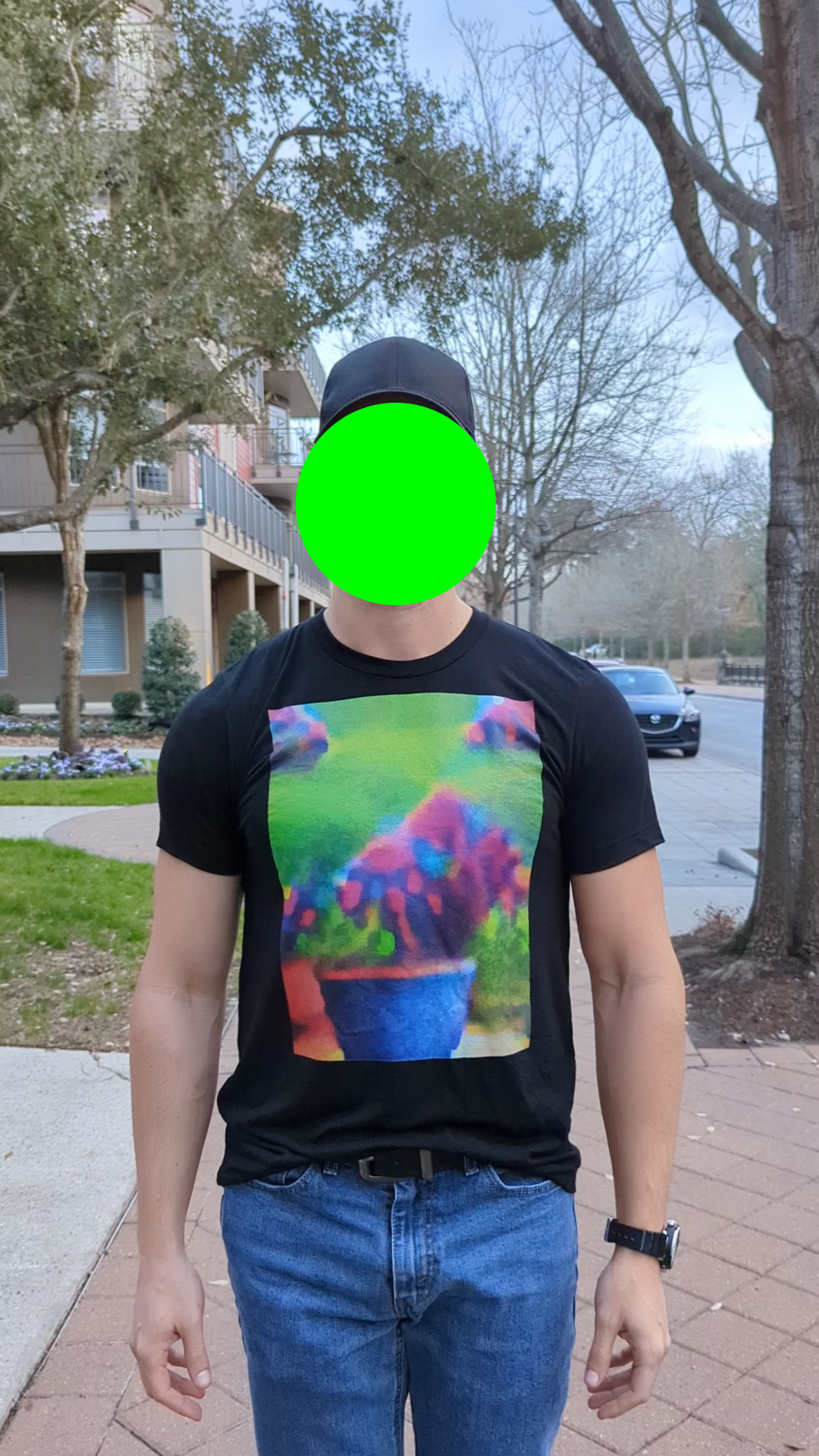}
                \caption{}
            \end{subfigure}
            \begin{subfigure}[b]{0.12\linewidth}
                \centering
                \includegraphics[width=\linewidth]{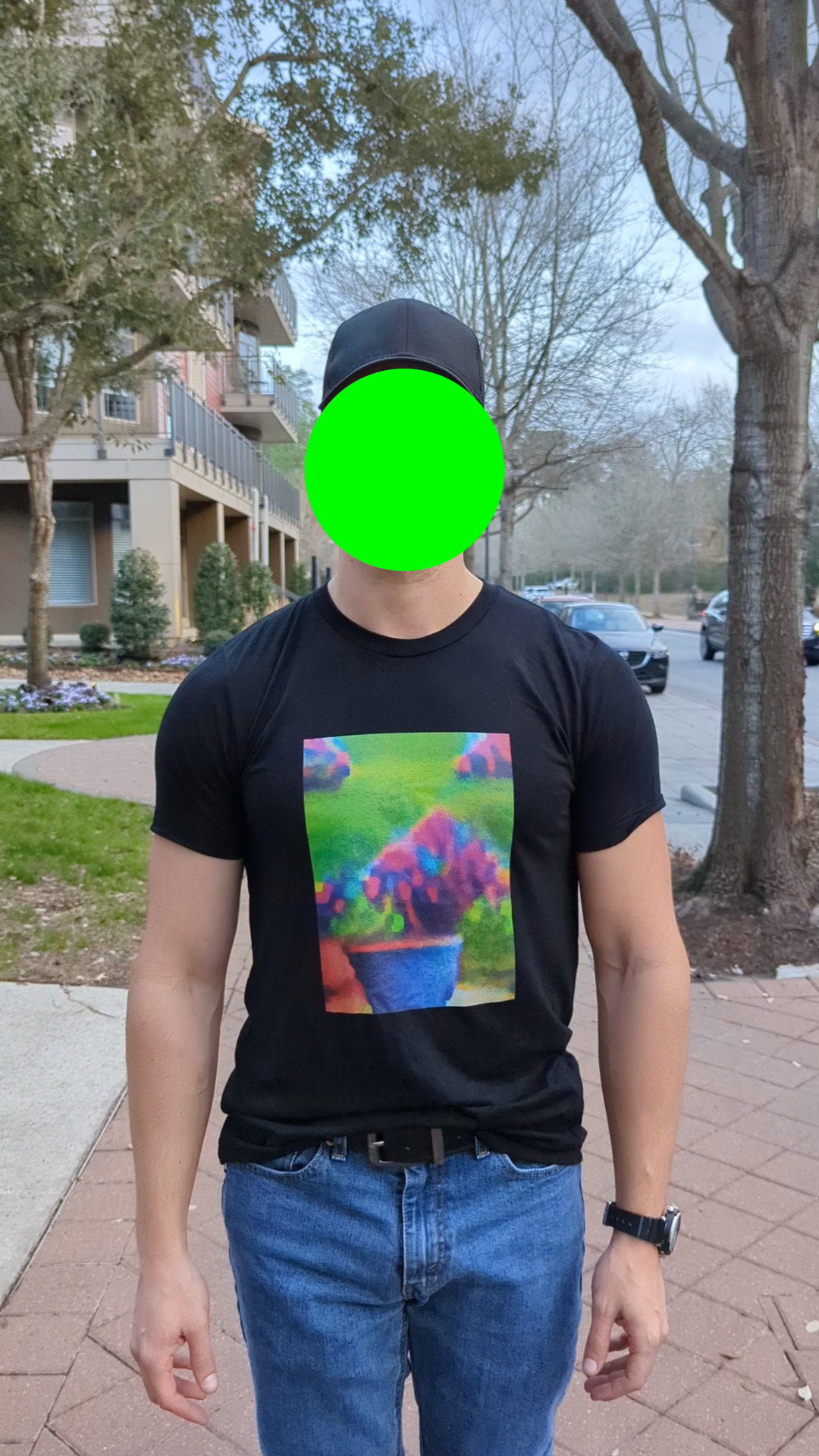}
                \caption{}
            \end{subfigure}
            \caption{Photographs of the four adversarial patches used for Experiment 1; (a) large-sized patch, trained at a large size (Standard); (b) small-sized patch, trained at a large size; (c) large-sized patch, trained at a small size; (d) small-sized patch, trained at a small size (Optimized)}
            \label{exp_1_patches}
        \end{figure}

        A dataset was constructed of 130 publicly available images of people in outdoor spaces, retrieved from Google Image Search by querying \textit{people outside}. These were deduplicated using ImageHash, keeping the largest of any duplicates \citep{buchner2016imagehash}. Images were annotated by the experimenters to include bounding boxes around each person, using LabelIMG \citep{tzutalin2015labelimg}.

        Two adversarial patches were generated by minimizing a combination of the model confidence for class person, the total variation of the patch, and color printability, following \citet{thys_fooling_2019}, against YOLOv2 \citet{redmon_you_2016}. The first patch was trained at 30\% the size of the total bounding box, which is approximately the size of the printable area that our vendor offers. We call this patch the "standard" patch. One shirt was printed with this patch covering the max area, and a second shirt was printed with a 50\% smaller patch.

        It was possible that the ASR of adversarial patches were influenced by their relative size during training, and that any size change would result in a reduced ASR. To control for this, a second patch was trained at 20\% the size of the bounding box, to be optimized for reduced patch sizes. We call this patch the "optimized" patch. Again, two t-shirts were printed: one with the patch covering the maximum printable area of the t-shirt, and a second that is smaller by 50\%.

        We were somewhat limited here in our vendor's inability to guarantee the size of the printed patches. The t-shirts we received had large size patches 35cm tall and 25cm wide. The reduced size patches were 25cm tall and 19cm wide. Examples of the four experimental patches are shown in Figure \ref{exp_1_patches}. For a control condition, a fifth shirt was printed, where the pattern on the front is a black-and-white checkerboard, following \citet{xu_evading_2019}.

        Videos were taken with a Google Pixel 7 of one participant wearing each shirt. During the video, the participant was instructed to turn their torso approximately 20 degrees to both the left and right, for a total recording time of approximately 6 seconds. A sidewalk in a residential area was used as the recording location. During testing the camera was placed at chest height, 2 meters away from the subject. These videos were sampled at 30fps using FFMPEG, resulting in approximately 200 test images per t-shirt \citep{tomar2006converting}.

        These samples were input to YOLOv2, running in GPU mode on an Nvidia Tesla K40, with both the detection threshold and non-max suppression (NMS) thresholds set to 40\%, following previous work. At each frame, the class label and prediction confidence of all detected objects were recorded. Frames were visually inspected to ensure that there was never a second person detected by the model. Any frame with no person detected by the model had an entry added for "person" with a confidence of $0.0$, to avoid biasing average statistics.

        Success was measured using ASR and the mean model confidence across samples. The motivation behind reporting the latter is that it's possible for two models to have a similar proportion of test samples below the detection threshold, but different impacts on the model above that threshold. Video processing, patch training, and inference were computed on Ubuntu Linux 18.04 using an Intel Core i5-4460 CPU, unless otherwise noted. Analysis was completed in Python using NumPy and Pandas \citep{harris2020array,mckinney-proc-scipy-2010}.

    \subsection{Results} \label{experiment_1_results}

        \begin{table}[h]
            \caption{Adversarial patch size results}
            \label{exp_1_results_table}
            \centering
            \begin{tabular}{c c}
                \toprule
                Patch Type & ASR (conf±95\%CI)\\
                \midrule
                Control & 0 (0.83±0.01) \\
                Large Size-optimized & 0 (0.70±0.01) \\
                Small Size-Optimized & 2 (0.64±0.02) \\
                Small Standard & 22 (0.50±0.03) \\
                Large Standard & \textbf{43} (\textbf{0.33±0.04}) \\
                \bottomrule
            \end{tabular}
        \end{table}

        Table \ref{exp_1_results_table} shows the results for each patch type. The most successful patch, with an ASR of 43\% is one which was trained at a large size, and deployed at a large size. The next most successful patch, which has just half the attack success of the first, was trained at a large size and deployed at a small size. The worst two performing patches are not appreciably better than the control condition when it comes to ASR, but succeed in reducing the average confidence.

        If the relative size during the training procedure is all that matters, then the optimized patch at the small size should perform better than the optimized patch at the large size, and similarly to the standard patch at the large size. On the other hand, if the total size is all that matters, we would expect to see the larger patches of either pattern type outperform smaller patches from either training procedure.

        In this case, we see a mixture of these results. The large standard patch performs best, but the large optimized patch is very close to the control condition. This may indicate that it's difficult to find gradients for very small patch sizes during the training.

\section{Experiment 2 - repeatable adversarial elements} \label{experiment_2}

    In experiment 2, we sought to construct a patch that is capable of covering the entire area of a shirt, to evaluate the effectivess of the largest possible patch. We were limited by our vendor's inability to guarantee position or orientation after constructing garments from shirts where the fabric itself is printed, and therefore sought to generate a pattern element that was small and relatively robust to changes in orientation.

    \subsection{Methods} \label{experiment_2_methods}

        \begin{figure}
            \centering
            \includegraphics*[width=0.5\linewidth]{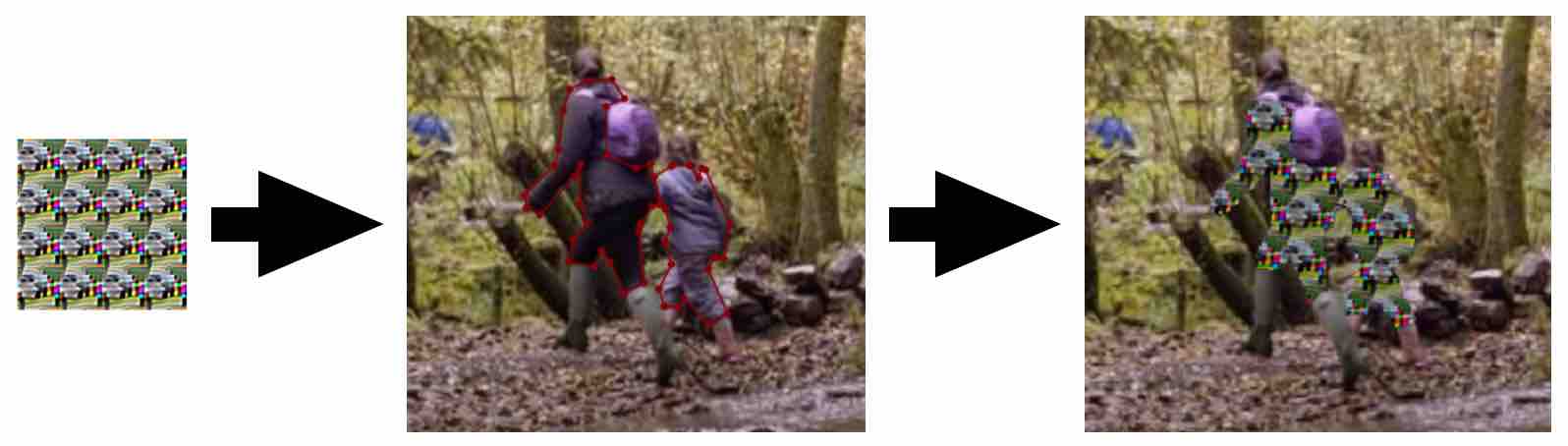}
            \caption{A depiction of the pattern training process, where a small element is repeated across the x and y extents of an image, then masked to only cover the portions of the image that contain printable materials like shirts or jackets.}
            \label{process}
        \end{figure}

        In experiment 2, we re-used the dataset from experiment 1. In addition to the existing YOLO bounding box labels, each image was re-annotated by the experimenters to include polygons around clothing in the photographs, using LabelMe \citep{Wada_Labelme_Image_Polygonal}.

        To generate an adversarial pattern composed of repetable elements, we start with a square patch as in \citet{thys_fooling_2019}, but resize it to be only 5\% the size of the bounding box in the training data. Using the mean box size creates variation during training with respect to the relative size of the element compared to the person. Additional heterogeneity is caused by persons not contained by the image, or who are unusually posed.

        Next, we perform a tiling operation, such that the patch is repeated in the x and y directions a sufficient number of times to be the same size as the entire input image. For instances where the input image is not evenly divisible by the patch size, the elements at the edges are truncated. Because we are not centering each pattern on each object, we introduce further patch heterogeneity in the training data.

        Finally, we use the clothing polygon annotations to create a boolean mask covering all of the clothing locations in the image, and only apply the tiled elements to the training image at true values. Because some persons in the training data are, e.g. wearing coats versus wearing tank tops, this adds variation for the total attack area. Tehse sources of heterogeneity can be though of as implicit additions to EOT. Figure \ref{process} shows a visual depiction of this process.

        A loss was calculated based on the reduction in maximum model confidence from the person class, along with the auxiliary losses reported earlier. During backpropagation, the loss for the patch in different locations in the image is combined into a single loss that is used to update the original, un-tiled patch.

        Test images were collected following the same procedure used for evaluation in \ref{experiment_1_methods}. The optimal printed elemtnt size of was determined by applying a pattern at different sizes to the training data, and observing which obtained the largest reduction in model confidence. The realized size of the elements on the t-shirt were 8cm by 8cm. The test pattern can be seen in Figure \ref{exp_2_patterns}a.

        During evaluation, the success of the adversarial pattern attack was evaluated under clean conditions and in the face of two adversarial defenses. First, we apply a resizing operation, reducing the image size by 50\% and 25\% before submitting it to the model. The original image sizes were 1920x1080, and were reduced to 960x540 and 480x270, respectively.

        Second, we use FFMPEG to convert the test images to the JPEG format. Initially we set the compressed image quality to 75\%, as \citet{das_keeping_2017} showed that this was sufficient to reduce ASR by 50\%. During testing this caused very little difference, so we added test evaluations at 50\% and 25\%.

    \subsection{Results} \label{experiment_2_results}
        \begin{table}[h]
            \caption{Reported attack success (YOLOv2)}
            \label{exp_2_results_table}
            \centering
            \begin{tabular}{c c c}
                \toprule
                Attack type & ASR (conf±95\%CI) \\
                \midrule
                Control & 0 (0.83±0.01) \\
                Adversarial patch (ours) & 42 (0.33±0.04) \\
                Adversarial t-shirt & 63 (-) \\
                Adversarial cloak & 50 (-) \\
                Adversarial pattern (ours) & \textbf{89} (\textbf{0.05±0.02}) \\
                \bottomrule
            \end{tabular}
        \end{table}

        In the no-defense case, we compare the ASR of our work against prior attacks against YOLOv2 (Table \ref{exp_2_results_table}). We omit for this comparison any virtual results, as these are known to be optimistic. Our training and realization methodology achieves state of the art attack success against YOLOv2.

        This should not be very surprising in light of the results presented in \ref{experiment_1_results}. Only the "Invisibility Cloak" paper includes an attack where the the adversarial patch is a similar size, and it is the closest to our own attack success. However, the improvement could be attributable to our evaluation method, and not the attack generation method. We will revisit this question in \ref{experiment_4_methods}.

        \begin{table}[h]
            \caption{Robustness to adversarial defenses}
            \label{exp_2_robustness_table}
            \centering
            \begin{tabular}{c c c}
                \toprule
                Defense type & ASR (conf±95\%CI) \\
                \midrule
                25\% resize & \textbf{7} (\textbf{0.49±0.02}) \\
                25\% quality jpg & 74 (0.13±0.03) \\
                50\% resize & 79 (0.11±0.02) \\
                50\% quality jpg & 82 (0.10±0.02) \\
                75\% quality jpg & 83 (0.84±0.02) \\
                No defense & 89 (0.05±0.02) \\
                \bottomrule
            \end{tabular}
        \end{table}

        For robustness testing, we observe that JPG compression does little to defend against adversarial patterns (Table \ref{exp_2_robustness_table}), where the most extreme JPG compression we attempted reduced our attack success only by 9 percentage points. The image resizing turned out to be much more effective. The 25\% reduction test scenario succeeded in reducing our ASR by 80 percentage points.

\section{Experiment 3 - attack transfer}
    In experiment 3, we sought to evaluate the success of the adversarial pattern in transfer scenarios. If an attack is trained against a known model, we want to evaluate how it performs against an unknown model. This is a compelling scenario since an attacker is unlikely to know the model architecture in the system under attack.

    \subsection{Methods} \label{experiment_3_methods}

\begin{figure}
            \centering
            \begin{subfigure}[b]{0.12\linewidth}
                \centering
                \includegraphics[width=\linewidth]{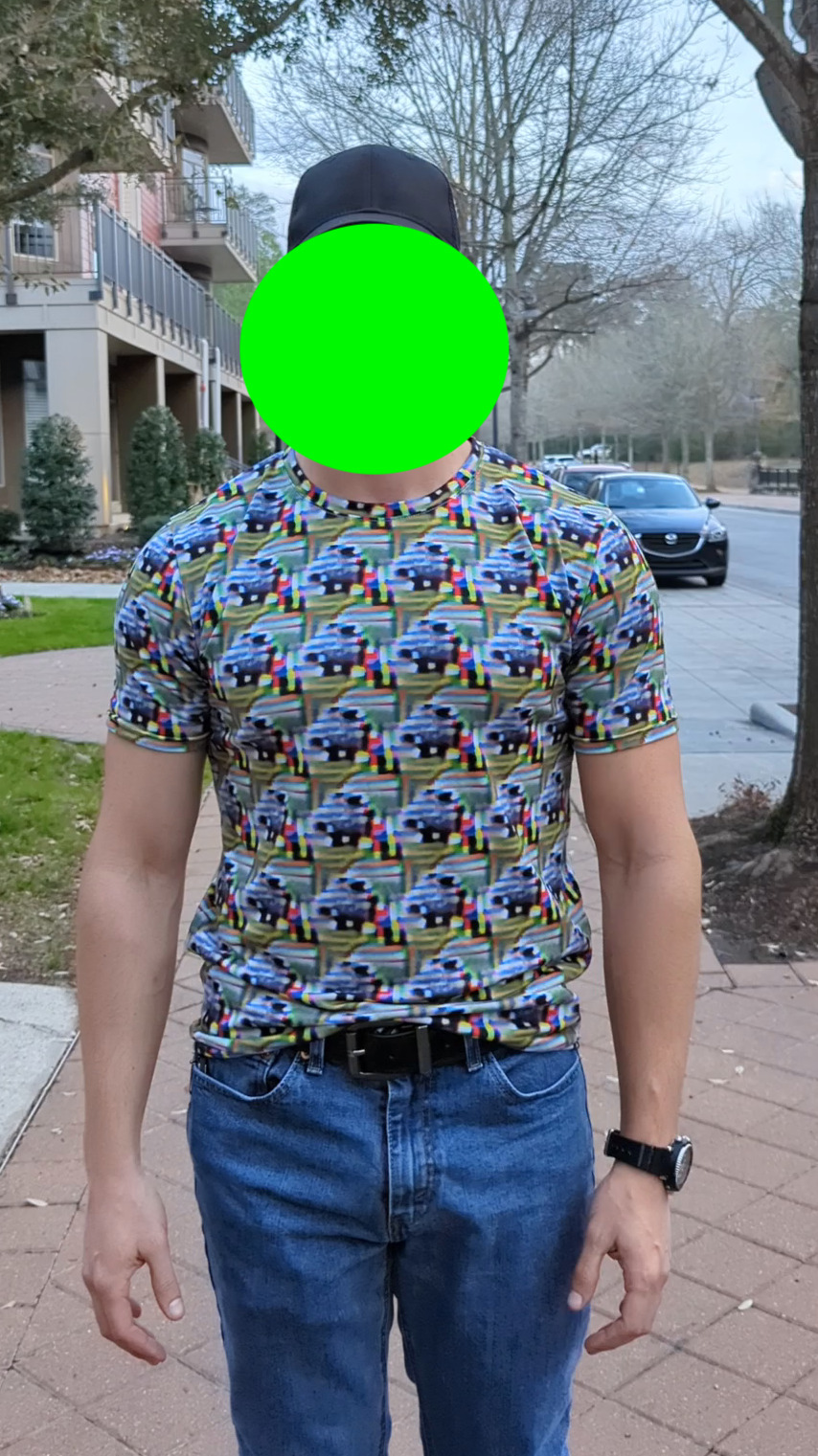}
                \caption{}
            \end{subfigure}
            \begin{subfigure}[b]{0.12\linewidth}
                \centering
                \includegraphics[width=\linewidth]{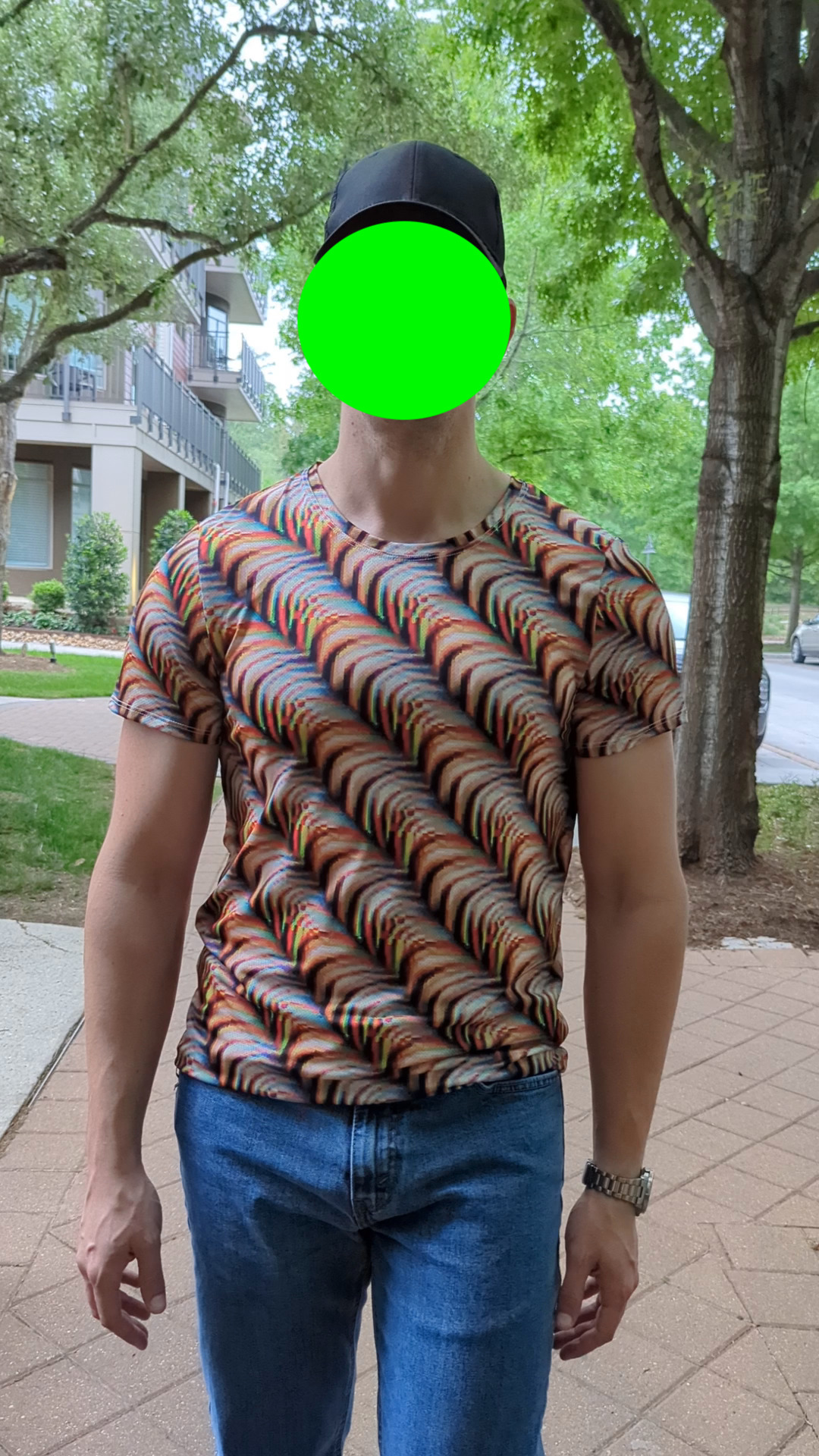}
                \caption{}
            \end{subfigure}
            \caption{Photographs of the two adversarial pattern attacks; (a) the YOLOv2 trained pattern; (b) the YOLOv3 trained pattern}
            \label{exp_2_patterns}
        \end{figure}

        For this experiment, we generate two sets of repeatable adversarial elements. For the first set, we re-use the pattern from \ref{experiment_2_methods}, which was trained against a YOLOv2 model. For the second, we train a new adversarial pattern, following the procedure from \ref{experiment_2_methods}, but using a YOLOv3 model running in GPU mode. The weights for this model were provided by \citet{redmon_yolov3_2018}. The optimal size for the YOLOv3 elements was a larger than the YOLOv2 attack, with each element measuring 9cm by 9cm. See figure \ref{exp_2_patterns} for examples.

        It was our intention to train a third attack using FasterRCNN, but were unable to compile the model repository against the versions of CUDA and CUDNN supported by our hardware. For evaluation, we used YOLOv2, YOLOv3, and Faster RCNN. YOLOv2 and v3 followed the same evaluation procedure as described in \ref{experiment_1_methods}, again with the detection and NMS thresholds set to 40\%.

        For Faster RCNN, we used the Resnet-50 C4 3xLR implementation, with weights provided by \citet{wu2019detectron2}. We set the ROI head threshold to 0.7, which was empirically determined to result in a single detection per object. Any test results with remaining duplicate detections used the maximum probability assigned to any detection, whose threshold was again set to 40\%.

        \subsection{Results} \label{experiment_3_results}
            \begin{table}[h]
                \caption{Measured attack transfer success}
                \label{exp_3_results_table}
                \centering
                \begin{tabular}{c c c c}
                    \multirow{2}{*}{Train model} & \multicolumn{3}{c}{Test model ASR (conf±95\%CI)}\\
                    & YOLOv2 & YOLOv3 & Faster RCNN \\
                    \midrule
                    YOLOv2 & \textbf{89} (\textbf{0.05±0.02}) & 0 (1.00±0.00) & 0 (0.87±0.01)\\
                    YOLOv3 & 53 (0.24±0.03) & \textbf{100} (\textbf{0.00±0.00}) & 0 (0.93±0.01)\\
                    \bottomrule
                \end{tabular}
            \end{table}

        The results from Experiment 3 are summarized in Table \ref{exp_3_results_table}. The YOLOv2 attack did not transfer to YOLOv3. The attack trained on YOLOv3 maintains some efficacy against YOLOv2, although its ASR is reduced by half. Neither of the YOLO model architecture trained attacks transferred to Faster RCNN.

        These results disagree with prior research arguing for robust attack transfer between models. It's possible that this is because articles like \citet{li_invisibilitee_2022}, \citet{huang_universal_2020}, and \citet{wu_making_2020} measured transfer success based on change in average multiclass precision and not in realized attack success. Attack success is a more difficult metric, since it only penalizes a score for failure to detect, and not for e.g. mistakes in position a bounding box.

        Indeed, Figure 8 from the "Invisbility Cloak" paper appears to show that training an attack on YOLOv2 has 0\% success when deployed against YOLOv3, although the authors do not comment on this \citep{wu_making_2020}. \citet{xu_evading_2019} noted that attack transfers tend to result in ASRs in the single digit percentages, which comports with our finding here.

\section{Experiment 4 - Replication}

    In \ref{experiment_2_results}, we compared the efficacy of our adversarial pattern against the reported success of attacks published previously in venues like CVPR and ECCV. In experiment 4, we repeated our test procedure from (\ref{experiment_2_methods}) on these adversarial patches to reduce any variance in results based on the testing procedure. We also hoped to validate that our own attack's high success rate was not due to an error in our testing procedure.

    \subsection{Methods} \label{experiment_4_methods}

        \begin{figure}
            \centering
            \begin{subfigure}[b]{0.16\linewidth}
                \centering
                \includegraphics[width=0.7\linewidth]{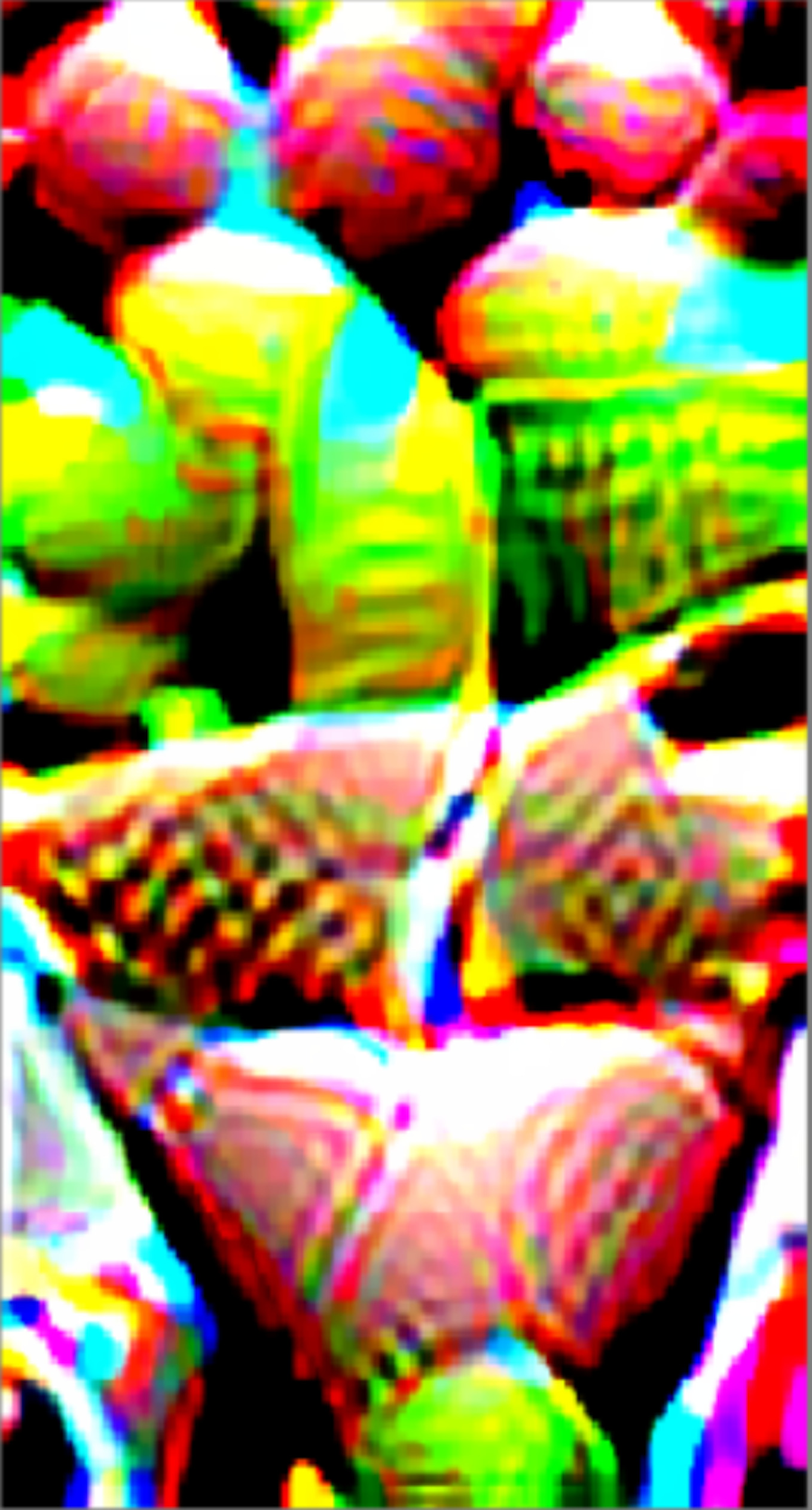}
                \caption{}
            \end{subfigure}
            \begin{subfigure}[b]{0.16\linewidth}
                \centering
                \includegraphics[width=0.75\linewidth]{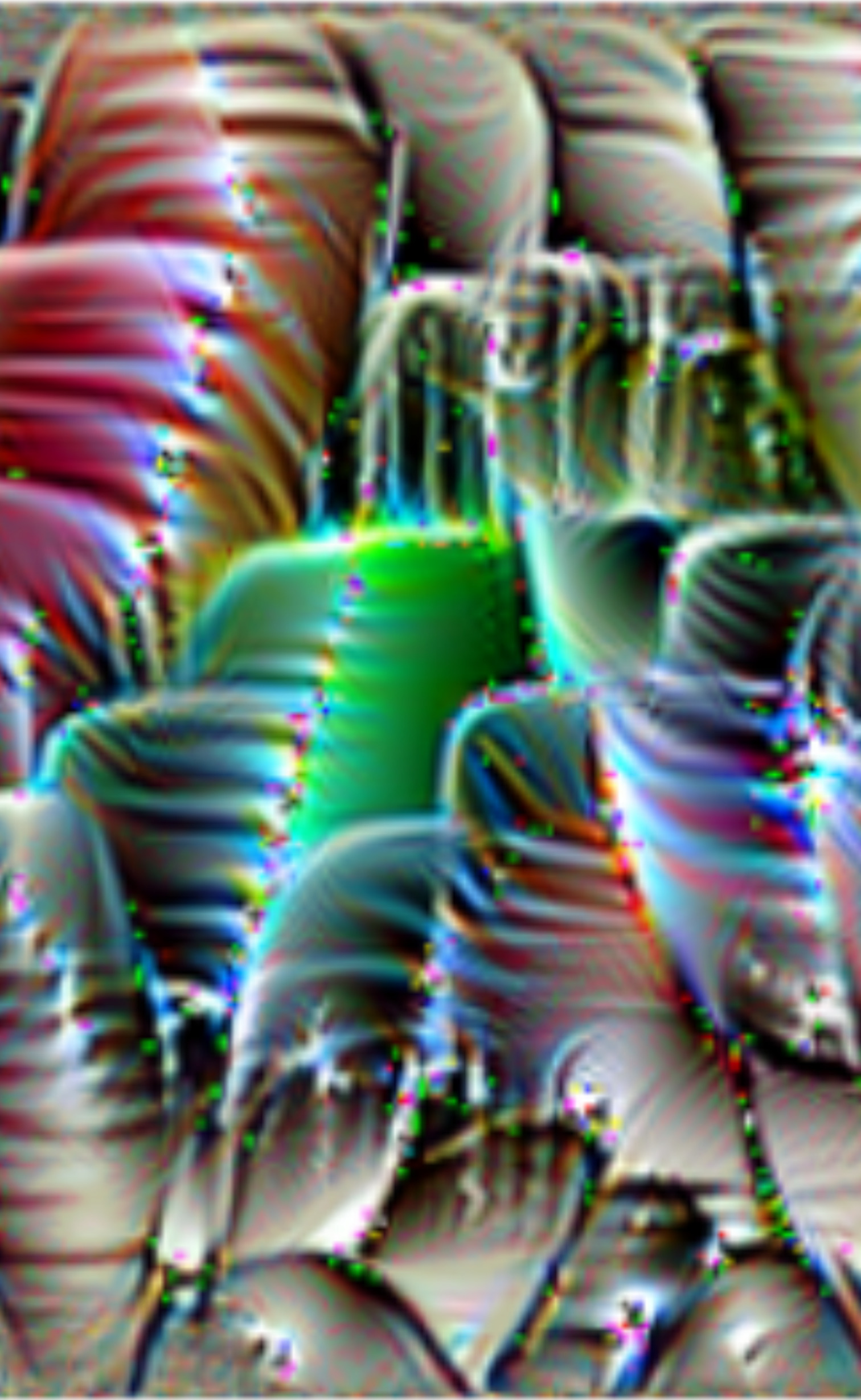}
                \caption{}
            \end{subfigure}
            \begin{subfigure}[b]{0.16\linewidth}
                \centering
                \includegraphics[width=0.74\linewidth]{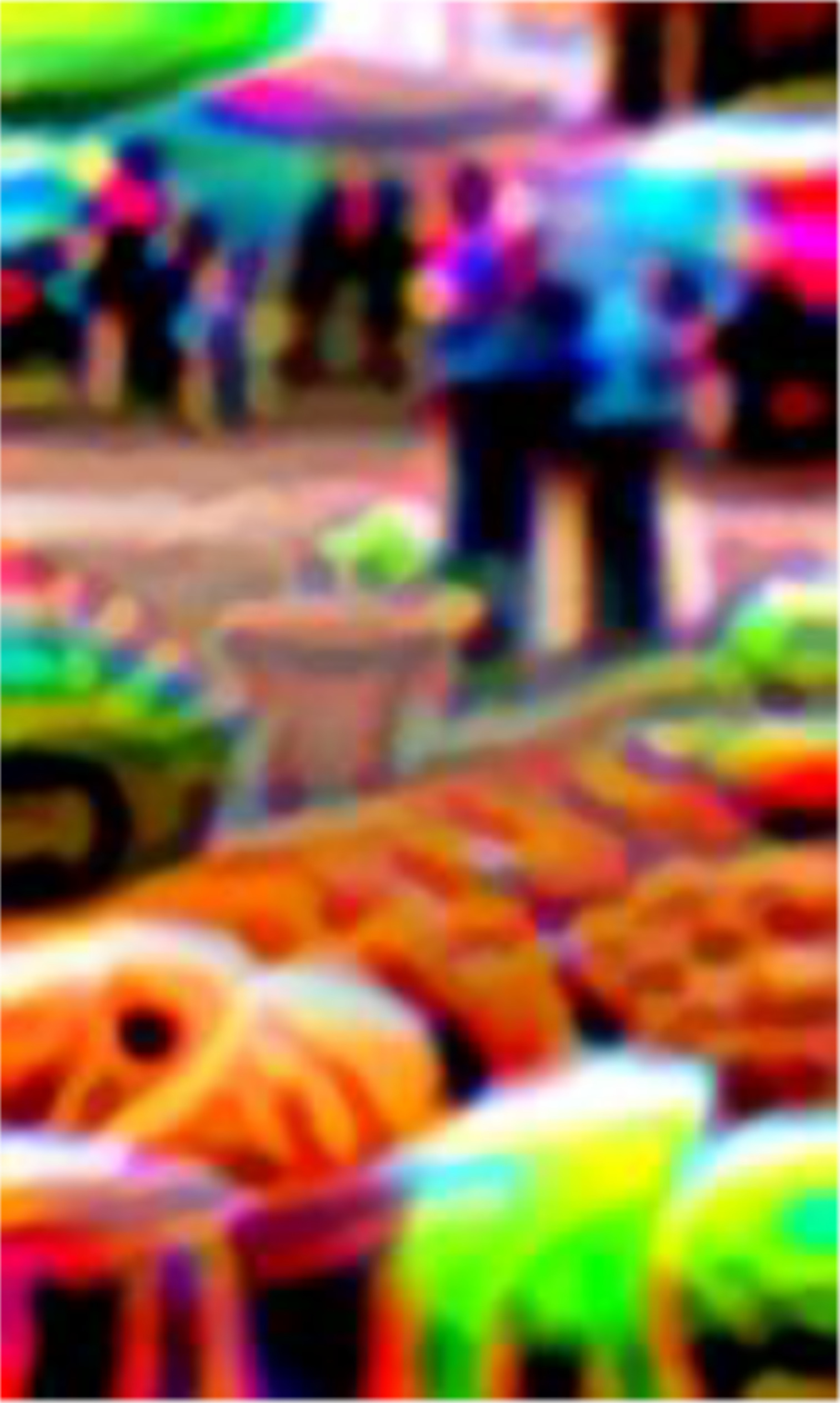}
                \caption{}
            \end{subfigure}
            \begin{subfigure}[b]{0.16\linewidth}
                \centering
                \includegraphics[width=0.75\linewidth]{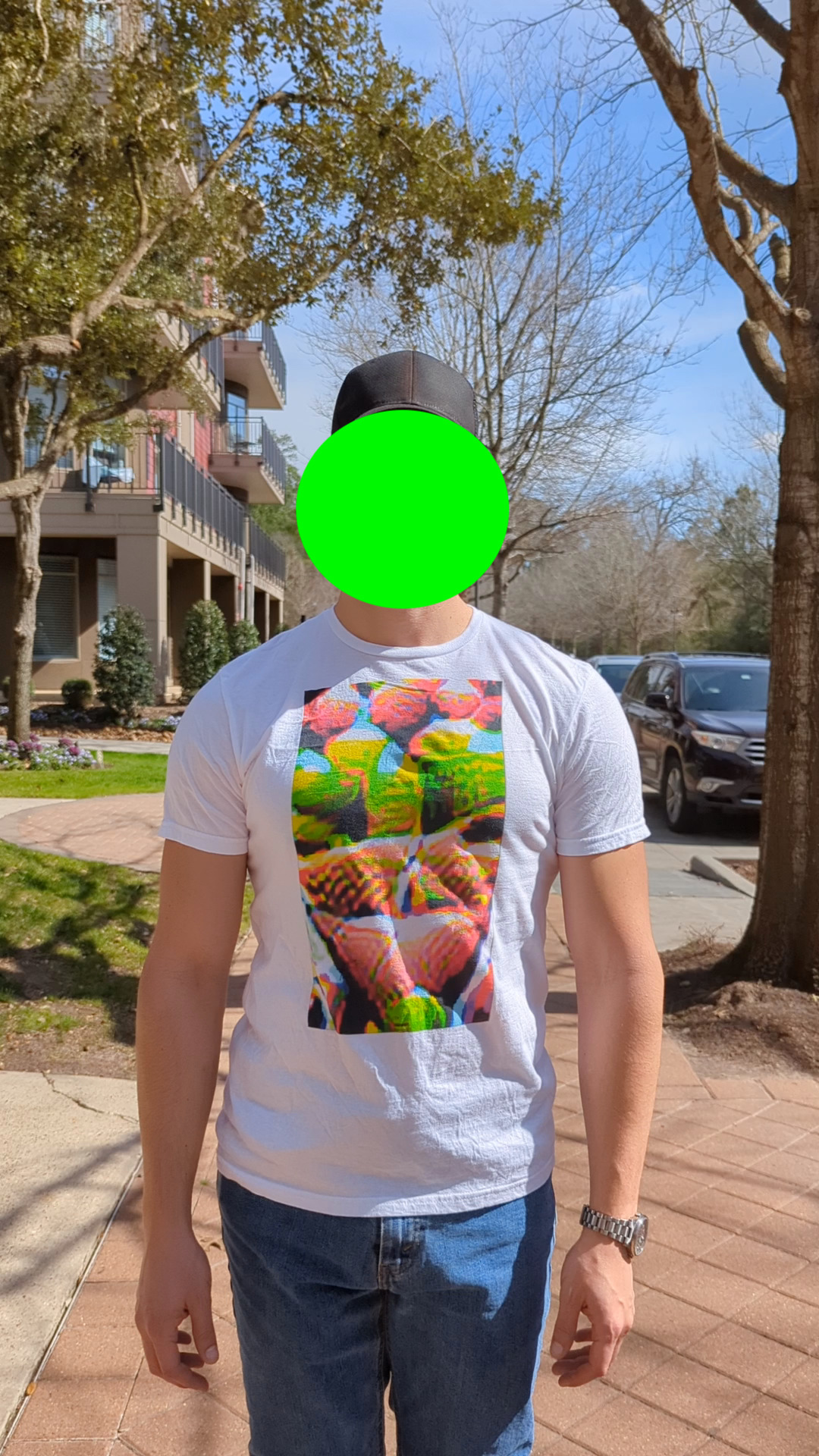}
                \caption{}
            \end{subfigure}
            \begin{subfigure}[b]{0.16\linewidth}
                \centering
                \includegraphics[width=0.75\linewidth]{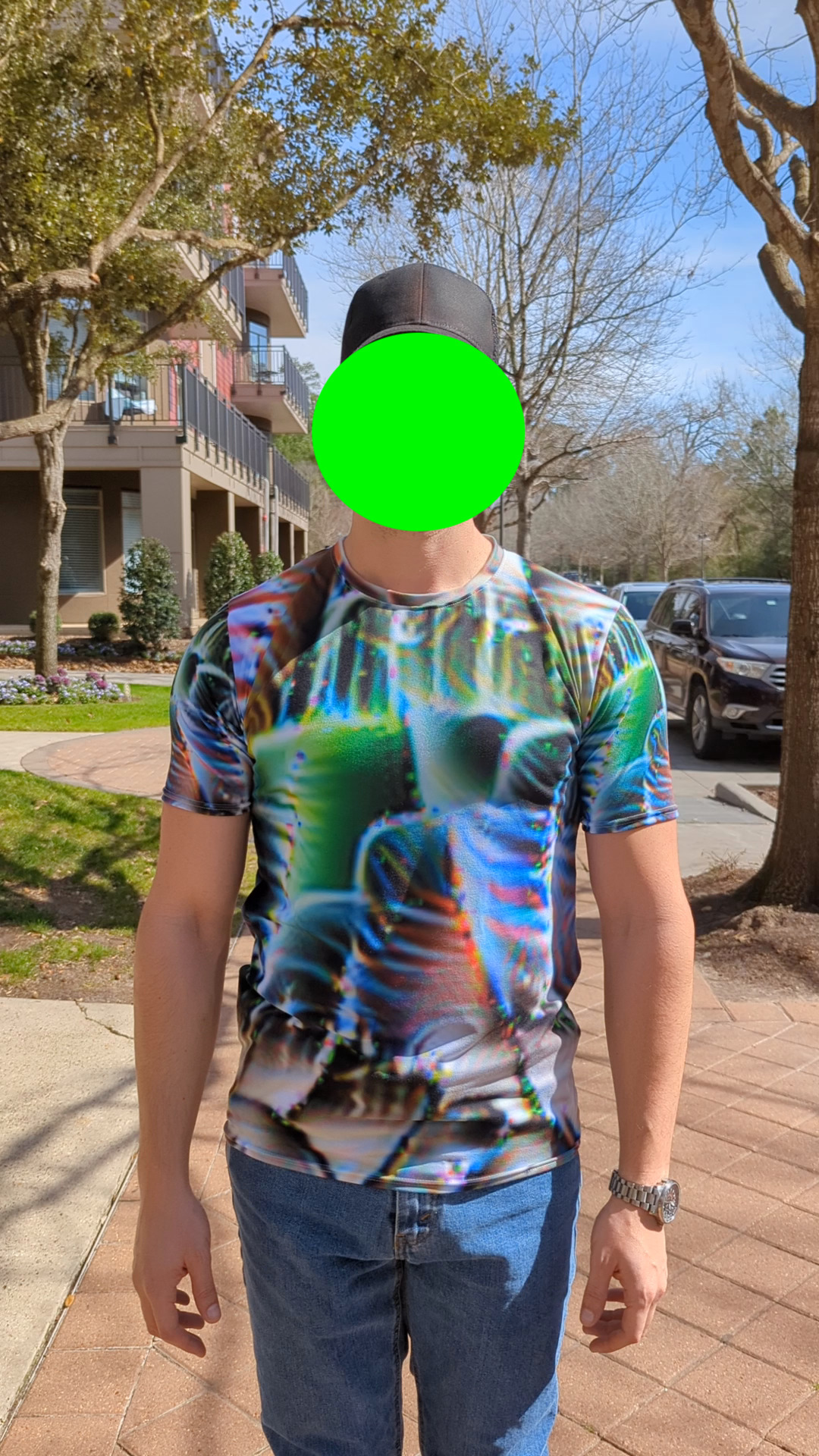}
                \caption{}
            \end{subfigure}
            \begin{subfigure}[b]{0.16\linewidth}
                \centering
                \includegraphics[width=0.75\linewidth]{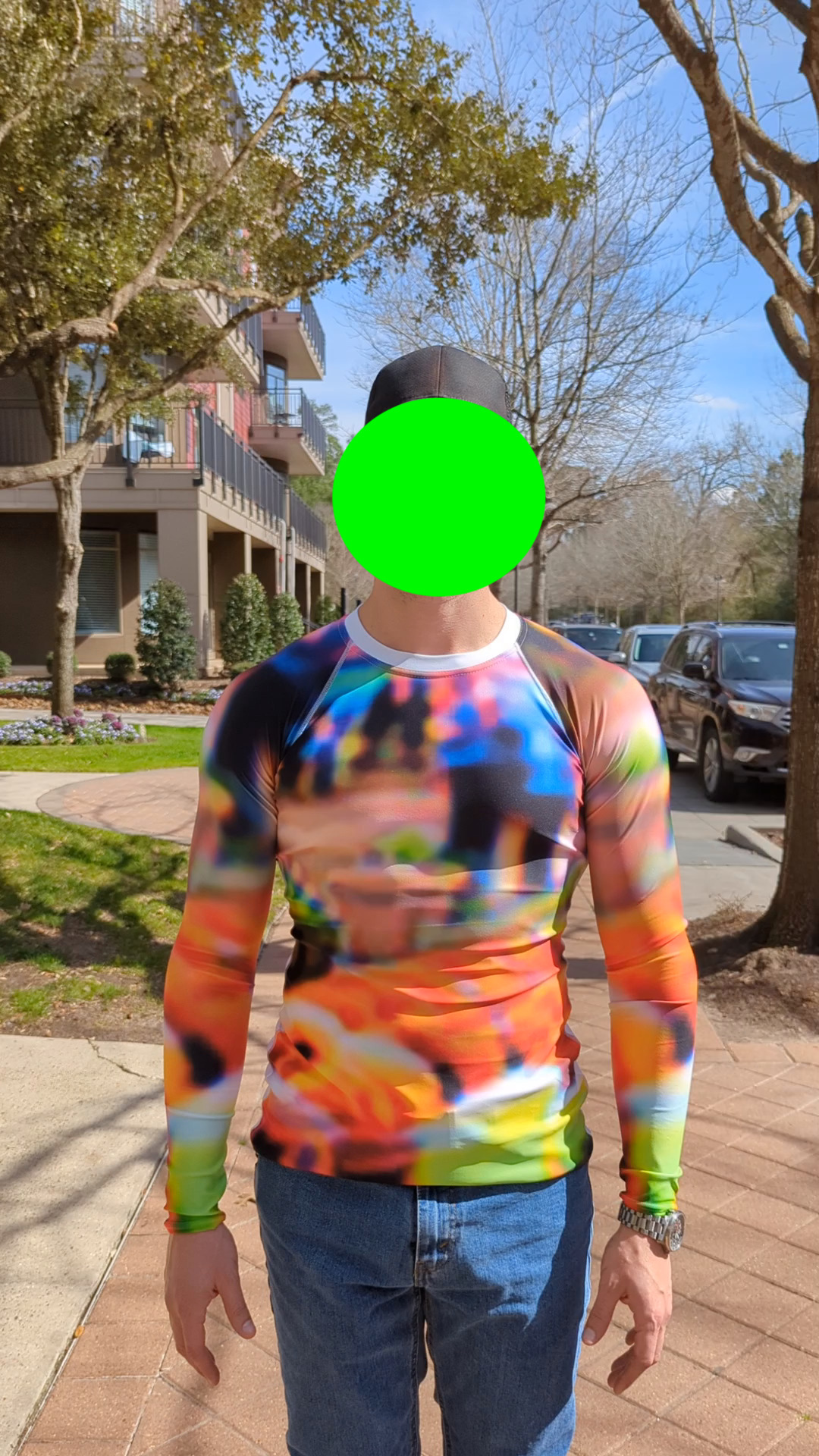}
                \caption{}
            \end{subfigure}
            \caption{Examples of the adversarial patch attacks that we attempt to reproduce: (a) "Adversarial T-shirt" patch; (b) "Invisibilitee" patch; (c) "Invisibility Cloak" patch; (d) "Adversarial T-shirt" realization; (e) "Invisibilitee" realization; (f) "Adversarial Cloak" realization}
            \label{exp_2_patches}
        \end{figure}

        To replicate attacks from prior art, we downloaded PDFs of three other adversarial clothing papers from arXiv: "adversarial t-shirt", "invisibilitee", and "invisbility cloak" \citep{xu_evading_2019,li_invisibilitee_2022,wu_making_2020}. In each paper, the figure depicting the adversarial patch was selected and the PDF was zoomed until the figure occupied the entire window. Then, a screenshot was taken of the patch. The selected patches can be seen in Figure \ref{exp_2_patches}a-c.

        For "adversarial t-shirt", a single rectangular patch was printed on the front of a white shirt, following \citet{xu_evading_2019}. The realized size of our patch was 38cm by 20cm, which appears smaller than their printed attack judging by the relative size of the patch compared to the size of the participant wearing the shirt.

        In "invisibilitee", \citet{li_invisibilitee_2022} present a number of successful patches. We selected the first one from Figure 7 and applied it to every fabric panel on a plain t-shirt. The authors did not test any physical realizations of this patch, so, lacking a better option, we centered the patch separately on each panel of fabric on the shirt.

        For "invisibility cloak", we used a long-sleeved shirt as in \citet{wu_making_2020}. We resized the patch until it covered the vertical extent of the shirt, and then mirrored it onto each sleeve. Examples of the realized attacks after printing can be seen in Figure \ref{exp_2_patches}d-f

        There are three potential sources of error in this methodology. The first is loss of fidelity in the transition from input figure, to screenshot, to printing. The second is an error in the size or orientation of the applied patch. The third is an error in how the patch was repeated to cover the available fabric. The success of a patch should be robust to these, considering the first two have been shown in our results above to have a measurable but non-debilitating impact on ASR.

        The evaluation methodology was identical to \ref{experiment_2_methods}. The attentive reader will have noticed in Figure \ref{exp_2_patches} that the lighting outside was different, due to a change in season.

    \subsection{Results} \label{experiment_4_results}

        \begin{table}[h]
            \caption{Replicated attack success}
            \label{exp_4_results_table}
            \centering
            \begin{tabular}{c c c c}
                \toprule
                \multirow{3}{*}{Attack type} & \multicolumn{3}{c}{ASR (conf±95\%CI)}\\
                & YOLOv2 & YOLOv3 & Faster RCNN \\
                \midrule
                Control & 0 (0.83±0.01) & 0 (1.00±0.00) & 0 (0.98±0.01)\\
                Adversarial T-shirt & 0 (0.75±0.01) & 0 (1.00±0.00) & 0 (0.95±0.01)\\
                Invisibility Cloak & 0 (0.84±0.01) & 0 (1.00±0.01) & 0 (\textbf{0.73±0.01})\\
                Invisibilitee & 0 (0.74±0.01) & 0 (1.00±0.00) & 0 (0.79±0.01)\\
                Adversarial Pattern (YOLOv2) & \textbf{89} (\textbf{0.05±0.02}) & 0 (0.99±0.00) & 0 (0.87±0.01)\\
                Adversarial Pattern (YOLOv3) & 53 (0.24±0.03) & \textbf{100} (\textbf{0.00±0.00}) & 0 (0.93±0.01)\\
                \bottomrule
            \end{tabular}
        \end{table}

        The results of Experiment 4 are summarized in Table \ref{exp_4_results_table}. We were unable to replicate the success of any of the attacks chosen for analysis. Not only were the success rates lower than reported, they were no better than the control condition.

        The average model confidences tell a more nuanced story. For Faster RCNN, "Invisbility Cloak" is the leader in terms of degrading model confidence with "Invisibilitee" as a close runner-up. The latter was trained against an implementation of Faster RCNN, which makes sense, but the former was trained on YOLOv2, which was surprising.

        For YOLOv2 and YOLOv3, the adversarial pattern attack presented in this paper has the highest success -- this is true whether you consider this replication attempt or the reported attack success (Table~\ref{prior_work_table}). For YOLOv2, the next best performer is the reported attack success of "Adversarial T-shirt" at 79\%. For YOLOv3, the next best attack would be the reported success of "Adversarial Cloak" at 40\%, but only the test result of using a large poster, and not the the printed sweater.

        There are a number of potential reasons why we were unable to replicate the success of previously published attacks. One possibility, which would agree with the data presented in \ref{experiment_3_results}, is that the sample attacks published in prior work were downsized. Assuming that an 80 percentage point reduction is typical for adversarial patches, this would result in reducing ASRs that had been in the range of 40\%-60\% all the way to zero.

        A second possibility is that our evaluation procedure differed significantly from that used in prior work. This includes a large number of potential factors, including taking test images outside, testing subjects that are close to the camera, or the resolution of that camera. It is also possible that our vendor cannot print images at the appropriate quality, which, given their difficulties printing squares at the correct size, would not be surprising.

\section{Discussion}

    \ref{experiment_1_results}, showed that reducing the size of an adversarial patch also reduced its effectiveness. Optimizing the patch to perform well at small sizes reverses this effect, with the larger version peforming worse than the smaller one. However, the size-optimized patch at any size performs worse than the standard patch at any size, indicating that the larger factor in explaining patch effectiveness is the total area.

    This interpretation would be supported by two other experimental results presented in this paper. First, in \ref{experiment_2_results}, we observed that the efficacy of adversarial patterns was appreciably better than patch-based attacks, but only marginally more effective than \citet{wu_making_2020}. We hypothesize that our gain in attack success was specifically a result of optimizing across the entire pattern, as opposed to optimizing just a patch, then marking it large post hoc. Second, our failure to reproduce the previously published success of attack methods could be attributed to image resizing performed during the compilation of conference papers, and not to problems with the patches themselves per se \ref{experiment_4_results}.

    \ref{experiment_2_results} showed that an adversarial pattern was not only capable of attacking an object detector, but that it achieved state of the art success. Because our approach can be used to cover any object that itself can be covered by a fabric or printable adhesive, this represents an opportunity to train attacks on large and irregularly shaped objects. There is prior art using differentiable approximations of 3D rendering engines to learn "adversarial textures" for airplanes \citet{salman_unadversarial_nodate}, but this extra computation may not be necessary.

    The pattern attack was robust to image compression algorithms, but not to image resizing. It's possible that resizing was smoothing over locally important information, and therefore destroying something critical for success. Another possibility is that FFMPEG uses an image resizing algorithm which gives different outputs than the bilinear approach used by Pillow \citep{clark2015pillow}.\footnote{See https://zuru.tech/blog/the-dangers-behind-image-resizing for a discussion of this problem in image preprocessing pipelines.}

    \ref{experiment_3_results}, showed that attacks on object detectors did not transfer well between iterations on model architectures, and not at all between different architectures. A likely interpretation of our experimental results and previously reported work is that the difference lies in the test metric used -- namely, that attack success is a more difficult metric.

    Finally, \ref{experiment_4_results} presented the results of directly replicating attacks published by other authors. To the best of our knowledge, this is the first paper to do so for attacks on object detectors. We were unable to replicate the success of any of the three papers we attempted, although as we discussed, there are a variety of reasons related to our methodology that could cause this.

    Having said that, we would still like to elevate points made previously by \citet{albert_ethical_2020} about testing methodologies for adversarial examples. The authors specifically are interested in the need to address diversity in testing contexts and participants, in particular where this relates to a reasonable approximation of a random sample, and not just the experimenter and their colleagues. We would like to note explicitly that our evaluation used just such a convenience sample of our colleagues.

    We would add to their list. To ensure the accuracy and generalizability of our test results, we should be deliberate about varying things like resolution, background, recording distance, and object color, as prior work has shown that DNNs are sensitive to these non-robust features \citep{hendrycks_natural_2019,ilyas_adversarial_2019,salman_adversarially_nodate}. This is especially important if we desire, as some authors do, to offer evasion attacks as a protection against the consequences of algorithms whose inner workings cannot be modified, and whose decisions cannot be contested \citep{kulynych_pots_2020}.

\section{Conclusion}

    Adversarial patches have garnered a lot of recent interest, in attacks as diverse as evading detection by CCTV cameras and conducting a "denial of service" attack against self-driving cars. We have provided evidence that the primary factor in attack success is the area covered by the patch itself. We further show the feasibility of adversarial patterns as an attack vector, which provides a new state of the art in evasion success against YOLOv2 and YOLOv3. We argue for more robust testing methodology for these kinds of attacks, given the interest in their use for both beneficient and harmful activities. We leave to future research the possibility of training adversarial patterns for non-person uses.

\bibliographystyle{abbrvnat}
\bibliography{paper}

\begin{thebibliography}{36}
\providecommand{\natexlab}[1]{#1}
\providecommand{\url}[1]{\texttt{#1}}
\expandafter\ifx\csname urlstyle\endcsname\relax
  \providecommand{\doi}[1]{doi: #1}\else
  \providecommand{\doi}{doi: \begingroup \urlstyle{rm}\Url}\fi

\bibitem[Albert et~al.(2020)Albert, Delano, Penney, Rigot, and
  Kumar]{albert_ethical_2020}
K.~Albert, M.~Delano, J.~Penney, A.~Rigot, and R.~S.~S. Kumar.
\newblock Ethical {Testing} in the {Real} {World}: {Evaluating} {Physical}
  {Testing} of {Adversarial} {Machine} {Learning}.
\newblock \emph{arXiv:2012.02048 [cs]}, Dec. 2020.
\newblock URL \url{http://arxiv.org/abs/2012.02048}.
\newblock arXiv: 2012.02048.

\bibitem[Athalye et~al.(2017)Athalye, Engstrom, Ilyas, and
  Kwok]{athalye_synthesizing_2017}
A.~Athalye, L.~Engstrom, A.~Ilyas, and K.~Kwok.
\newblock Synthesizing {Robust} {Adversarial} {Examples}.
\newblock \emph{arXiv:1707.07397 [cs]}, July 2017.
\newblock URL \url{http://arxiv.org/abs/1707.07397}.
\newblock arXiv: 1707.07397.

\bibitem[Biggio and Roli(2018)]{biggio_wild_2018}
B.~Biggio and F.~Roli.
\newblock Wild {Patterns}: {Ten} {Years} {After} the {Rise} of {Adversarial}
  {Machine} {Learning}.
\newblock \emph{Pattern Recognition}, 84:\penalty0 317--331, Dec. 2018.
\newblock ISSN 00313203.
\newblock \doi{10.1016/j.patcog.2018.07.023}.
\newblock URL \url{http://arxiv.org/abs/1712.03141}.
\newblock arXiv:1712.03141 [cs].

\bibitem[Brown et~al.(2017)Brown, Mané, Roy, Abadi, and
  Gilmer]{brown_adversarial_2017}
T.~B. Brown, D.~Mané, A.~Roy, M.~Abadi, and J.~Gilmer.
\newblock Adversarial {Patch}.
\newblock \emph{arXiv:1712.09665 [cs]}, Dec. 2017.
\newblock URL \url{http://arxiv.org/abs/1712.09665}.
\newblock arXiv: 1712.09665.

\bibitem[Buchner(2016)]{buchner2016imagehash}
J.~Buchner.
\newblock Imagehash.
\newblock \url{https://github.com/JohannesBuchner/imagehash}, 2016.

\bibitem[Clark(2015)]{clark2015pillow}
A.~Clark.
\newblock Pillow (pil fork) documentation, 2015.
\newblock URL
  \url{https://buildmedia.readthedocs.org/media/pdf/pillow/latest/pillow.pdf}.

\bibitem[Das et~al.(2017)Das, Shanbhogue, Chen, Hohman, Chen, Kounavis, and
  Chau]{das_keeping_2017}
N.~Das, M.~Shanbhogue, S.-T. Chen, F.~Hohman, L.~Chen, M.~E. Kounavis, and
  D.~H. Chau.
\newblock Keeping the {Bad} {Guys} {Out}: {Protecting} and {Vaccinating} {Deep}
  {Learning} with {JPEG} {Compression}, May 2017.
\newblock URL \url{http://arxiv.org/abs/1705.02900}.
\newblock arXiv:1705.02900 [cs].

\bibitem[Das()]{das_subversive_nodate}
S.~Das.
\newblock Subversive {AI}: {Resisting} automated algorithmic surveillance with
  human-centered adversarial machine learning.
\newblock page~4.

\bibitem[Everingham et~al.(2010)Everingham, Van~Gool, Williams, Winn, and
  Zisserman]{everingham_pascal_2010}
M.~Everingham, L.~Van~Gool, C.~K.~I. Williams, J.~Winn, and A.~Zisserman.
\newblock The {Pascal} {Visual} {Object} {Classes} ({VOC}) {Challenge}.
\newblock \emph{International Journal of Computer Vision}, 88\penalty0
  (2):\penalty0 303--338, June 2010.
\newblock ISSN 0920-5691, 1573-1405.
\newblock \doi{10.1007/s11263-009-0275-4}.
\newblock URL \url{http://link.springer.com/10.1007/s11263-009-0275-4}.

\bibitem[Eykholt et~al.(2017)Eykholt, Evtimov, Fernandes, Li, Rahmati, Xiao,
  Prakash, Kohno, and Song]{eykholt_robust_2017}
K.~Eykholt, I.~Evtimov, E.~Fernandes, B.~Li, A.~Rahmati, C.~Xiao, A.~Prakash,
  T.~Kohno, and D.~Song.
\newblock Robust {Physical}-{World} {Attacks} on {Deep} {Learning} {Models}.
\newblock \emph{arXiv:1707.08945 [cs]}, July 2017.
\newblock URL \url{http://arxiv.org/abs/1707.08945}.
\newblock arXiv: 1707.08945.

\bibitem[Girshick et~al.(2014)Girshick, Donahue, Darrell, and
  Malik]{girshick_rich_2014}
R.~Girshick, J.~Donahue, T.~Darrell, and J.~Malik.
\newblock Rich feature hierarchies for accurate object detection and semantic
  segmentation, Oct. 2014.
\newblock URL \url{http://arxiv.org/abs/1311.2524}.
\newblock arXiv:1311.2524 [cs].

\bibitem[Goodfellow et~al.(2014)Goodfellow, Shlens, and
  Szegedy]{goodfellow_explaining_2014}
I.~J. Goodfellow, J.~Shlens, and C.~Szegedy.
\newblock Explaining and {Harnessing} {Adversarial} {Examples}.
\newblock \emph{arXiv:1412.6572 [cs, stat]}, Dec. 2014.
\newblock URL \url{http://arxiv.org/abs/1412.6572}.
\newblock arXiv: 1412.6572.

\bibitem[Harris et~al.(2020)Harris, Millman, van~der Walt, Gommers, Virtanen,
  Cournapeau, Wieser, Taylor, Berg, Smith, Kern, Picus, Hoyer, van Kerkwijk,
  Brett, Haldane, del R{\'{i}}o, Wiebe, Peterson, G{\'{e}}rard-Marchant,
  Sheppard, Reddy, Weckesser, Abbasi, Gohlke, and Oliphant]{harris2020array}
C.~R. Harris, K.~J. Millman, S.~J. van~der Walt, R.~Gommers, P.~Virtanen,
  D.~Cournapeau, E.~Wieser, J.~Taylor, S.~Berg, N.~J. Smith, R.~Kern, M.~Picus,
  S.~Hoyer, M.~H. van Kerkwijk, M.~Brett, A.~Haldane, J.~F. del R{\'{i}}o,
  M.~Wiebe, P.~Peterson, P.~G{\'{e}}rard-Marchant, K.~Sheppard, T.~Reddy,
  W.~Weckesser, H.~Abbasi, C.~Gohlke, and T.~E. Oliphant.
\newblock Array programming with {NumPy}.
\newblock \emph{Nature}, 585\penalty0 (7825):\penalty0 357--362, Sept. 2020.
\newblock \doi{10.1038/s41586-020-2649-2}.
\newblock URL \url{https://doi.org/10.1038/s41586-020-2649-2}.

\bibitem[Hendrycks et~al.(2019)Hendrycks, Zhao, Basart, Steinhardt, and
  Song]{hendrycks_natural_2019}
D.~Hendrycks, K.~Zhao, S.~Basart, J.~Steinhardt, and D.~Song.
\newblock Natural {Adversarial} {Examples}.
\newblock \emph{arXiv:1907.07174 [cs, stat]}, July 2019.
\newblock URL \url{http://arxiv.org/abs/1907.07174}.
\newblock arXiv: 1907.07174.

\bibitem[Huang et~al.(2020)Huang, Gao, Zhou, Xie, Yuille, Zou, and
  Liu]{huang_universal_2020}
L.~Huang, C.~Gao, Y.~Zhou, C.~Xie, A.~Yuille, C.~Zou, and N.~Liu.
\newblock Universal {Physical} {Camouflage} {Attacks} on {Object} {Detectors},
  Apr. 2020.
\newblock URL \url{http://arxiv.org/abs/1909.04326}.
\newblock arXiv:1909.04326 [cs].

\bibitem[Ilyas et~al.(2019)Ilyas, Santurkar, Tsipras, Engstrom, Tran, and
  Madry]{ilyas_adversarial_2019}
A.~Ilyas, S.~Santurkar, D.~Tsipras, L.~Engstrom, B.~Tran, and A.~Madry.
\newblock Adversarial {Examples} {Are} {Not} {Bugs}, {They} {Are} {Features}.
\newblock \emph{arXiv:1905.02175 [cs, stat]}, May 2019.
\newblock URL \url{http://arxiv.org/abs/1905.02175}.
\newblock arXiv: 1905.02175.

\bibitem[Krizhevsky et~al.(2017)Krizhevsky, Sutskever, and
  Hinton]{krizhevsky_imagenet_2017}
A.~Krizhevsky, I.~Sutskever, and G.~E. Hinton.
\newblock {ImageNet} classification with deep convolutional neural networks.
\newblock \emph{Communications of the ACM}, 60\penalty0 (6):\penalty0 84--90,
  May 2017.
\newblock ISSN 0001-0782, 1557-7317.
\newblock \doi{10.1145/3065386}.
\newblock URL \url{https://dl.acm.org/doi/10.1145/3065386}.

\bibitem[Kulynych et~al.(2020)Kulynych, Overdorf, Troncoso, and
  Gürses]{kulynych_pots_2020}
B.~Kulynych, R.~Overdorf, C.~Troncoso, and S.~Gürses.
\newblock {POTs}: protective optimization technologies.
\newblock pages 177--188. ACM, Jan. 2020.
\newblock ISBN 978-1-4503-6936-7.
\newblock \doi{10.1145/3351095.3372853}.
\newblock URL \url{https://dl.acm.org/doi/10.1145/3351095.3372853}.

\bibitem[Li et~al.(2022)Li, Zhang, Zhao, Zhang, Liu, Wang, and
  Wen]{li_invisibilitee_2022}
Y.~Li, B.~Zhang, G.~Zhao, M.~Zhang, J.~Liu, Z.~Wang, and J.~Wen.
\newblock {InvisibiliTee}: {Angle}-agnostic {Cloaking} from {Person}-{Tracking}
  {Systems} with a {Tee}, Aug. 2022.
\newblock URL \url{http://arxiv.org/abs/2208.06962}.
\newblock Number: arXiv:2208.06962 arXiv:2208.06962 [cs].

\bibitem[Lin et~al.(2015)Lin, Maire, Belongie, Bourdev, Girshick, Hays, Perona,
  Ramanan, Zitnick, and Dollár]{lin_microsoft_2015}
T.-Y. Lin, M.~Maire, S.~Belongie, L.~Bourdev, R.~Girshick, J.~Hays, P.~Perona,
  D.~Ramanan, C.~L. Zitnick, and P.~Dollár.
\newblock Microsoft {COCO}: {Common} {Objects} in {Context}, Feb. 2015.
\newblock URL \url{http://arxiv.org/abs/1405.0312}.
\newblock arXiv:1405.0312 [cs].

\bibitem[Liu et~al.(2019)Liu, Yang, Liu, Song, Li, and Chen]{liu_dpatch_2019}
X.~Liu, H.~Yang, Z.~Liu, L.~Song, H.~Li, and Y.~Chen.
\newblock {DPatch}: {An} {Adversarial} {Patch} {Attack} on {Object}
  {Detectors}, Apr. 2019.
\newblock URL \url{http://arxiv.org/abs/1806.02299}.
\newblock arXiv:1806.02299 [cs].

\bibitem[Redmon and Farhadi(2016)]{redmon_yolo9000_2016}
J.~Redmon and A.~Farhadi.
\newblock {YOLO9000}: {Better}, {Faster}, {Stronger}.
\newblock \emph{arXiv:1612.08242 [cs]}, Dec. 2016.
\newblock URL \url{http://arxiv.org/abs/1612.08242}.
\newblock arXiv: 1612.08242.

\bibitem[Redmon and Farhadi(2018)]{redmon_yolov3_2018}
J.~Redmon and A.~Farhadi.
\newblock {YOLOv3}: {An} {Incremental} {Improvement}, Apr. 2018.
\newblock URL \url{http://arxiv.org/abs/1804.02767}.
\newblock arXiv:1804.02767 [cs].

\bibitem[Redmon et~al.(2016)Redmon, Divvala, Girshick, and
  Farhadi]{redmon_you_2016}
J.~Redmon, S.~Divvala, R.~Girshick, and A.~Farhadi.
\newblock You {Only} {Look} {Once}: {Unified}, {Real}-{Time} {Object}
  {Detection}.
\newblock \emph{arXiv:1506.02640 [cs]}, May 2016.
\newblock URL \url{http://arxiv.org/abs/1506.02640}.
\newblock arXiv: 1506.02640.

\bibitem[Ren et~al.(2017)Ren, He, Girshick, and Sun]{ren_faster_2017}
S.~Ren, K.~He, R.~Girshick, and J.~Sun.
\newblock Faster r-cnn: Towards real-time object detection with region proposal
  networks.
\newblock \emph{IEEE Transactions on Pattern Analysis and Machine
  Intelligence}, 39\penalty0 (6):\penalty0 1137--1149, 2017.
\newblock \doi{10.1109/TPAMI.2016.2577031}.

\bibitem[Salman et~al.({\natexlab{a}})Salman, Ilyas, Engstrom, Kapoor, and
  Ma]{salman_adversarially_nodate}
H.~Salman, A.~Ilyas, L.~Engstrom, A.~Kapoor, and A.~Ma.
\newblock Do {Adversarially} {Robust} {ImageNet} {Models} {Transfer} {Better}?
\newblock page~13, {\natexlab{a}}.

\bibitem[Salman et~al.({\natexlab{b}})Salman, Vemprala, Ilyas, Ma, Engstrom,
  and Kapoor]{salman_unadversarial_nodate}
H.~Salman, S.~Vemprala, A.~Ilyas, A.~Ma, L.~Engstrom, and A.~Kapoor.
\newblock Unadversarial {Examples}: {Designing} {Objects} for {Robust}
  {Vision}.
\newblock {\natexlab{b}}.

\bibitem[Szegedy et~al.(2013)Szegedy, Zaremba, Sutskever, Bruna, Erhan,
  Goodfellow, and Fergus]{szegedy_intriguing_2013}
C.~Szegedy, W.~Zaremba, I.~Sutskever, J.~Bruna, D.~Erhan, I.~Goodfellow, and
  R.~Fergus.
\newblock Intriguing properties of neural networks.
\newblock \emph{arXiv:1312.6199 [cs]}, Dec. 2013.
\newblock URL \url{http://arxiv.org/abs/1312.6199}.
\newblock arXiv: 1312.6199.

\bibitem[Thys et~al.(2019)Thys, Van~Ranst, and Goedemé]{thys_fooling_2019}
S.~Thys, W.~Van~Ranst, and T.~Goedemé.
\newblock Fooling automated surveillance cameras: adversarial patches to attack
  person detection.
\newblock \emph{arXiv:1904.08653 [cs]}, Apr. 2019.
\newblock URL \url{http://arxiv.org/abs/1904.08653}.
\newblock arXiv: 1904.08653.

\bibitem[Tomar(2006)]{tomar2006converting}
S.~Tomar.
\newblock Converting video formats with ffmpeg.
\newblock \emph{Linux Journal}, 2006\penalty0 (146):\penalty0 10, 2006.

\bibitem[Tzutalin(2015)]{tzutalin2015labelimg}
Tzutalin.
\newblock Labelimg.
\newblock \url{https://github.com/tzutalin/labelImg}, 2015.

\bibitem[Wada()]{Wada_Labelme_Image_Polygonal}
K.~Wada.
\newblock {Labelme: Image Polygonal Annotation with Python}.
\newblock URL \url{https://github.com/wkentaro/labelme}.

\bibitem[{W}es {M}c{K}inney(2010)]{mckinney-proc-scipy-2010}
{W}es {M}c{K}inney.
\newblock {D}ata {S}tructures for {S}tatistical {C}omputing in {P}ython.
\newblock In {S}t\'efan van~der {W}alt and {J}arrod {M}illman, editors,
  \emph{{P}roceedings of the 9th {P}ython in {S}cience {C}onference}, pages 56
  -- 61, 2010.
\newblock \doi{10.25080/Majora-92bf1922-00a}.

\bibitem[Wu et~al.(2019)Wu, Kirillov, Massa, Lo, and
  Girshick]{wu2019detectron2}
Y.~Wu, A.~Kirillov, F.~Massa, W.-Y. Lo, and R.~Girshick.
\newblock Detectron2.
\newblock \url{https://github.com/facebookresearch/detectron2}, 2019.

\bibitem[Wu et~al.(2020)Wu, Lim, Davis, and Goldstein]{wu_making_2020}
Z.~Wu, S.-N. Lim, L.~Davis, and T.~Goldstein.
\newblock Making an {Invisibility} {Cloak}: {Real} {World} {Adversarial}
  {Attacks} on {Object} {Detectors}, July 2020.
\newblock URL \url{http://arxiv.org/abs/1910.14667}.
\newblock Number: arXiv:1910.14667 arXiv:1910.14667 [cs, math].

\bibitem[Xu et~al.(2019)Xu, Zhang, Liu, Fan, Sun, Chen, Chen, Wang, and
  Lin]{xu_evading_2019}
K.~Xu, G.~Zhang, S.~Liu, Q.~Fan, M.~Sun, H.~Chen, P.-Y. Chen, Y.~Wang, and
  X.~Lin.
\newblock Evading {Real}-{Time} {Person} {Detectors} by {Adversarial}
  {T}-shirt.
\newblock \emph{arXiv:1910.11099 [cs]}, Oct. 2019.
\newblock URL \url{http://arxiv.org/abs/1910.11099}.
\newblock arXiv: 1910.11099.

\end{thebibliography}

\end{document}